\documentclass{article}
\usepackage{ijcai19}

\usepackage{times}
\usepackage{soul}
\usepackage{url}
\usepackage[hidelinks]{hyperref}
\usepackage[utf8]{inputenc}
\usepackage[small]{caption}
\usepackage{graphicx}
\usepackage{amsmath}
\usepackage{booktabs}
\usepackage{algorithm2e}
\usepackage{array}
\usepackage{graphicx}
\usepackage{adjustbox}
\graphicspath{ {./images/} }

\makeatletter
\def\blfootnote{\xdef\@thefnmark{}\@footnotetext}
\makeatother

\title{Generating Counterfactual and Contrastive Explanations using SHAP}

\author{
    Shubham Rathi
    \affiliations
    IIIT Hyderabad, India \emails
    shubham.rathi@research.iiit.ac.in
}

\begin{document}

\maketitle

\begin{abstract}
  With the advent of GDPR, the domain of explainable AI and model interpretability has gained added impetus. Methods to extract and communicate visibility into decision-making models have become legal requirement. Two specific types of explanations, contrastive and counterfactual have been identified as suitable for human understanding. In this paper, we propose a model agnostic method and its systemic implementation to generate these explanations using shapely additive explanations (SHAP). We discuss a generative pipeline to create contrastive explanations and use it to further to generate counterfactual datapoints. This pipeline is tested and discussed on the IRIS, Wine Quality \& Mobile Features dataset. Analysis of the results obtained follows. 
\end{abstract}

\section{Introduction}

AI often has a critique that when effective, it is a black box \cite{Castelvecchi}. This black box problem of AI inhibits its adoption in critical systems where algorithmic decisions can have social impact on people. Prior to EU's General Data Protection Regulation (GDPR) on the `Right to Explanation', efforts to create explainable AI (xAI) was driven by the need to understand, optimize and enhance the performance of complex models, build trust with the user and control the autonomy of machines \cite{Gunning}. GDPR has tilted the equation from a know-how to a legal necessity for any systems having algorithmic decision making. 
This paper and its work is intended to contribute in making systems GDPR compliant by systematizing a model agnostic method of generating explanations for instances where algorithms have discretion (classification and recommendation tasks) and thereby abet in making more accountable, transparent and thus explainable AI. In Section I, we introduce the topic and the nature of work so far. In Section II, we discuss our approach and results of using Shapley Additive Explanations \cite{shap} to generate contrastive and counterfactual explanations and when applied in a systemic setting. The paper concludes in Section III with a discussion on future improvements in the problem space. \blfootnote{* This work was presented at 2nd Workshop on Humanizing AI (HAI) at IJCAI'19 in Macao, China}

\subsection{Background \& Motivation}
\subsubsection{GDPR's Right to Explanation}
General Data Protection Regulation (GDPR) is a regulation focused on data protection and regulations regarding algorithmic decision-making and is abiding on companies operating in the European Union. One of the controversial regulations of this directive is the `Right to Explanation' which allows those significantly (socially) impacted by the decision of an algorithm to demand an explanation or rationale behind the decision (Eg: Being denied a loan application). This is especially a challenging ask given that many of the effective machine learning models (complex Neural Nets) are more or less black boxes even to its developers. Also, disclosing the inner workings of proprietary models might expose trade secrets and make the systems vulnerable to gamification. In such a scenario, giving a consistent and legally viable explanation to the end user is a challenge. Though this directive is not legally binding in all scenarios, it is very much a footprint about the course of future legislation in this area given increased awareness about user privacy and added engagement with autonomous computational entities. Since this regulation, there is increased research in making platforms GDPR compliant and generating explanations which allow the end user to rationalize the decision process. 
\par So far, there are minimal working implementations of a system or conversational AI agents that generate explanations. One such example is of Hendricks et all \cite{hendriks} wherein they use an explanation model
to predict candidate counterfactual evidence on images, or evidence which is discriminative for a counter-class. It is then verified if counterfactual evidence is in a given image using an evidence checker. Having access to sentences which describe what is in an image, to generate counterfactual explanations, the corresponding phrases are negated to generate a cohesive counterfactual explanation. A major downside of this method is that the dataset has to be annotated which might not be feasible in most practical cases. Also, this approach is not model agnostic and so is applicable only in similar usecases. 
Jasper van der Waa et all attempt to create congruent contrastive explanations using Foil Trees. The method utilizes locally trained one-versus-all decision trees to identify the disjoint set of rules that causes the tree to classify data points as the foil and not as the fact \cite{jasper}. A downside of their approach is that they use LIME to generate the decision tree and hence samples in the vicinity of the particular data point receive higher weights in training the tree. Which is, the decision tree is biased towards the neighbourhood. 
In another work, these researchers also attempt at generating Contrastive Explanations for Reinforcement Learning in terms of Expected Consequences \cite{jasper2}. Sokol et al. \cite{sokol} generate class contrastive counterfactual explanations for decision trees and logical machine learning models such as rule lists. Their approach takes advantage of access to the internal structure of a decision tree to measure pairwise distance between all its leafs. This approach has a downside that it relies heavily on the decision structure of the data-structure (decision tree, rule list) and hence is not truly model agnostic.  Our approach is an improvement over the downsides of the previous methods: our approach is model agnostic and also globally more consistent because of the use of Shapley Additive explanations (SHAP). Details about SHAP have been discussed in the next section.
\par While there are not many working solutions to the problem we've stated, there is much research at the task of generating contrastive and counterfactual explanations which has been elaborated in the section on related work. 
\subsubsection{Explanations \& its typology}
In this paper, we generate `contrastive' and `counterfactual' explanations. There are many types of explanations and much study has been devoted to understand what kind of explanations will be useful for humans. Thus, besides Computer Scientists, the GDPR directive has opened the domain of explainable AI to lawyers, philosophers, regulators and ethicists making this a classic multidisciplinary problem in the narrow context of GDPR. 
\par Building the problem ground up, it begins with the focus on a very specific aspect: What is an explanation and what sort of explanation is useful for humans? Falling back on decades of debates in philosophy and computer science, there have come to become some typologies of explanations. \\
Depending on their completeness or degree to which the entire causal chain be explained \cite{ruben}, they can be categorized as:  
\begin{itemize}
  \item Partial: address why particular facts occurred \cite{miller}
  \item Scientific: explain the general scientific relationships \cite{miller} between factors.
\end{itemize}
Depending on the model behavior, explanations can also be:
\begin{itemize}
  \item Post-Hoc (after this): a post-mortem manner of generating explanation formulated purely on the basis of model behavior generated after the occurrence of the predicted event. 
  \item Ante-Hoc (before this): These explanations seek to understand the inner working of a model while the model is in the process of making decisions. 
\end{itemize}
If explanations are to be constructed for humans, they are required to be contrastive, selective and socially interactive \cite{Mittelstadt}. They can be:
\begin{itemize}
  \item Contrastive: of the form `Why P not Q?' (alternative question) or `Why P but Q' (congruent question).  From the perspective of artificial intelligence, the former is asking why a particular algorithm gave an output rather than some other output that the questioner expected, while the latter is asking why an algorithm gave a particular output this time but some (probably different) output another time \cite{lipton}. Lipton's research concludes that if explanations are to be designed for humans, they should be contrastive \cite{lipton}. There has been study on the typology of alternative questions by Van Bouwel and Weber \cite{bouwel}:
  \begin{itemize}
     \item  P-contrast:  Why does object a have property P, rather than property Q?
     \item  O-contrast:  Why does object a have property P, while object b has property Q?
     \item  T-contrast:  Why does object a have property P at time t, but property Q at time t?
  \end{itemize}
  \item Counterfactual: of the form `If P then Q' or statement of how the world would have to be different for a desirable outcome to occur. To put it simply, a counterfactual explanation is the minimum possible change required to generate the desired output. Multiple counterfactuals might exist as multiple desirable outcomes can exist, and there may be several ways to achieve any of these outcomes \cite{sandra}. In their publication \cite{sandra}, Wachter et all have argued that Counterfactual explanations are GDPR compliant.
  \end{itemize}
  As seen, philosophy has vigorously dealt with the notion of explanation and thus a shallow understanding of the concept might be vague and misleading. The purpose of introducing this typology and nomenclature is to sensitize the reader on the nature of our work. In this paper, we demonstrate how \textit{Partial Posthoc P-type Contrastive explanations} and corresponding Counterfactual explanation (data points) can be generated using SHAP. 

\subsection{Related Work}
\subsubsection{Scientific Modelling}
In the field of interpretable machine learning (iML), there have been substantial efforts in getting posthoc insights into models. As of today, there are about 84 distinct iML methods \cite{robeer}. All methods can either be Global or Local. Global approaches aim to explain the complete model. Strictly speaking, they require an explanation in which the explainee is able to comprehend an aspect of the entire model at once \cite{zachary}. Local models only seek to explain a single decision by the neighborhood around the data point it predicted, and can therefore sometimes disregard large parts of the model in their explanation \cite{edwards}. Thus, local approximations are accurate representations only of a specific `slice' of a model \cite{Mittelstadt}. A popular local model approximation technique is Local Interpretable Model-agnostic Explanations (LIME) \cite{LIME}. According to the paper, LIME is `an algorithm that can explain the predictions of any classifier or regressor in a faithful way, by approximating it locally with an interpretable model'. SHAP (SHapley Additive exPlanations) \cite{shap} is a game theory based glocal additive feature attribution method to explain the output of any model and is the basis of our explanandum. 

\subsubsection{Generating explanandum}
There is substantial research at generating Contrastive explanations mostly using local models. Sandra et all \cite{sandra} first expounded the notion of Counterfactual explanations as aptly suited for GDPR and proposed an optimization equation for the same. The basic idea of counterfactual is that a counterfactual should be as similar as possible to the instance regarding feature values with change of as few features as possible \cite{molnar}. This definition is the foci of our proposed method using Shapley values. Watcher's technique of generating counterfactual points is so far only theoretical and involves defining a loss function that takes as input the instance of interest, a counterfactual and the desired (counterfactual) outcome. The loss measures how far the predicted outcome of the counterfactual is from to the predefined outcome and how far the counterfactual is from the instance of interest  \cite{molnar}. 

\section{Counterfactual and Contrastive Explanations using SHAP}
As stated earlier, this paper attempts to generate Partial Posthoc P-type Contrastive explanations and the corresponding Counterfactual datapoints. 
Our explanations are partial because the intent is to generate explanations that can be fathomed by humans. The complexity of models can be put out in scientific and mathematical ways but that would defeat the purpose of explaining. The authors have previously worked on an ontological approach on generating ex-ante explanations \cite{iswc}, this paper takes the stride to generate explanations in a post-hoc manner. The explanations generated in our methodology are P-Contrastive because we allow the user queries of the format `Why P not Q?'. In addition to Contrastive explanations, we also provide with Counterfactual datapoints which align with the contrastive explanations to enable the user visibility into the change in datum necessary to achieve a specific output. 
\subsection{Shapley Additive Explanations (SHAP)} 
SHAP is a is a unified approach to explain the output of any machine learning model recently developed by S Lundberg et all \cite{shap}. SHAP connects game theory with local explanations. SHAP values come with the black box local estimation advantages of LIME, but also come with theoretical guarantees about consistency and local accuracy from game theory. The difference between the prediction and the average prediction is fairly distributed among the features values of the instance and is the shapley efficiency property. This property sets the Shapley value apart from other methods like LIME. LIME does not guarantee to perfectly distribute the effects. It might make the Shapley value the only method to deliver a full explanation \cite{molnar}. The basic tradeoff between SHAP and LIME is that LIME does not offer a globally consistent explanation while SHAP does. SHAP has been developed and released as a python toolset for iML wherein corresponding to each feature, SHAP returns a list of Shapley values for a specific datum. This is based on the idea that predictions can be explained by assuming that each feature is a `player' in a game where the prediction is the payout. The Shapley value - a method from coalitional game theory - tells us how to fairly distribute the ‘payout’ among the features \cite{molnar}. The interpretation of the Shapley value $\phi \textsubscript{ij}$ for feature j and instance i is: the feature value x\textsubscript{ij} contributed $\phi \textsubscript{ij}$ towards the prediction for instance i compared to the average prediction for the dataset \cite{molnar}. In our approach, we use Shapley values to determine which factors work for or against a particular classification. 

\subsection{Approach}
Given any datapoint, a classifier predicts its given output class. Keeping the current datapoint as reference, a P-contrast question is of the format `Why [predicted-class] not [desired-class]?'. By specifying the desired class, we limit our search space to a single alternative from the multiple alternatives otherwise. Given the datapoint, we estimate its Shapley values for each of the possible target classes. The negative Shapley values indicate the features that have negatively contributed to the specific class classification and vice-versa. Thus, to generate Natural Language explanations for questions like `Why P not Q', we break down the answer in two segments: `Why P?' and `Why not Q?'. The answer for these two segments is constructed using the Shapley values for class P and class Q and returned. Treading the definition of Counterfactuals \cite{molnar}, we mutate only the features that work against the classification of the desired category and achieve the counterfactual datapoints. These data points are a counterfactual answer to the user's contrastive query. A description of the approach to generate the Natural Language explanation and the Contrastive datapoint is given below. It begins by the generation of a list of Shapley Values using the SHAP package \footnote{https://github.com/slundberg/shap} corresponding to each of the classes. Note that this approach can also work on continuous datapoints but for the explanation, we assume discreet classes. 

\begin{algorithm}
\SetAlgoLined
\KwData {$dp=Input()$ is the datapoint,\\ $Q=Input()$ is the desired class $\neq$ P.}
\KwResult{Shapley values for a given datapoint generated from the SHAP toolset }
$P \longleftarrow $\ Classifier($dp$), 
$Q \longleftarrow $\ $Q$,
$SV \longleftarrow $\ SHAP($dp$)\\
\KwRet{$P, Q, SV$}
 \caption{Find P, Q \& Shapley Values}
\end{algorithm}

The pipeline begins by identifying the desired class (Q), the predicted class (P) and the data point. Shapley values are generated for each of the target classes which are further used to generate the contrastive and counterfactual explanations.

\begin{algorithm}
\SetAlgoLined
\KwData {P, Q, SV}
\KwResult{Contrastive explanation to `Why P not Q?' }
$Positive \longleftarrow $\ ($SV[P]$) $>$ 0\\ 
$Negative \longleftarrow $\ ($SV[Q]$) $<$ 0\\
$whyP \longleftarrow $\ $generateNLExp(Positive)$ \\ 
$notQ \longleftarrow $\ $generateNLExp(Negative)$\\
\KwRet{$whyP, notQ$}
 \caption{Generate Contrastive Explanation}
\end{algorithm}

The generated Shapley values are used to generate explanations in Natural language. This same explanation could also be expressed as a histogram. 

\begin{algorithm}
\SetAlgoLined
\KwData {$SV$, $dp$, $Q$ }
\KwResult{Counterfactual Datapoints}
 \For{$i\leftarrow 1$ \KwTo $length(Training Datapoints)$}{
 $counterFactuals \leftarrow$ $None$ \\
 $noOfPoints \leftarrow$ 50 * $i$ \\
 $MutateFeatures \leftarrow$ $SV[Q] < 0$ \\
 \For{$point$ \KwTo $NearestNeighbours(noOfPoints)$}{
 $mutatedDatapoint \leftarrow$ $dp$ \\
 $mutatedDatapoint[MutatedFeatures] \leftarrow$ $point[MutatedFeatures]$ \\
 \If{$Classify(mutatedDatapoint)$ == $Q$}{
   Add $mutatedDatapoint$ to $counterFactuals$\;
   }
 }
 \If{$length(counterFactuals) >$ $0$}{
   \KwRet{$counterFactuals$};
   }
 }
 \KwRet{$None$}
 \caption{Generate Counterfactual Datapoints}
\end{algorithm}

Finally, we use the Shapley values to also generate Counterfactual explanations. We begin by generating nearest neighbors in multiples of 50. If counterfactuals are found in these points, we return else we continue. As mentioned earlier, we only mutate those features which contribute against the classification of Q. This pipeline is primarily based on the efficacy of SHAP and to the best of our understanding, is the only globally consistent method of generating contrastive and counterfactual explanations since this approach is the first extension of the use of Shapley Additive explanations to this problem space.

\subsection{Results}
\subsubsection{Generation of Explanation and Counterfactual points}
Our methodology was tested on three datasets: The mobile feature dataset \footnote{https://www.kaggle.com/iabhishekofficial/mobile-price-classification}, the IRIS dataset, the Wine Quality Dataset \footnote{https://archive.ics.uci.edu/ml/datasets/wine+quality} and could be extended for other datasets. For a data point, the system generated a contrastive explanation and counterfactual datapoints from the method discussed above. For instance, for a data point from the IRIS dataset, the following results were obtained for the query `Why 0 not 1' : \\

\begin{tabular}{|c|c|}
\hline 
Original Datapoint
& \multicolumn{1}{|p{4.3cm}|}{\centering [4.4, 2.9, 1.4, 0.2]}\\
\hline 
\hline 
Counterfactual points
& \multicolumn{1}{|p{4.3cm}|}{\centering [4.4, 2.9, 1.4, 0.2],\par [4.4, 2.9, 3.0, 0.2],\par [4.4, 2.9, 3.3, 0.2],\par [4.4, 2.9, 3.5, 0.2],\par [4.4, 2.9, 3.7, 0.2],\par [4.4, 2.9, 3.8, 0.2],\par [4.4, 2.9, 3.9, 0.2],\par [4.4, 2.9, 4.0, 0.2]}\\
\hline
\hline 
Why 0?
& \multicolumn{1}{|p{4.3cm}|}{ Algorithms Pro classification was primarily influenced by petal width (cm)}\\
\hline 
\hline 
Why not 1?
& \multicolumn{1}{|p{4.3cm}|}{ Algorithms Anti classification was primarily influenced by petal length (cm)}\\
\hline 
\end{tabular} \\

We compare the process of generation of these counterfactual points on various models. The models tested are K-Nearest Neighbour (KNN), Neural Network (NN), Random Forest (RF) \& Support Vector Machine (SVM). The following section shows the result on various datasets and its analysis.  

\subsubsection{Results on datasets}

As mentioned, we have tested our pipeline on the IRIS, Wine Quality and the Mobile features dataset. On each of these datasets, we apply K-Nearest Neighbour, Random Forest, A simple LBFGS Neural Network and Linear kernel SVM. Our evaluation criteria is different from that of other research in this area. The closest comparison to our work is that of Jasper van der Waa et al \cite{jasper} who measure the efficacy of their trained decision tree and its fidelity with the underlying model using measures such as F\textsubscript{1} Scores (Accuracy), Mean length of explanation and the time required to generate the explanation. We instead measure the efficacy of our approach by the total number of novel counterfactual points (points which are not in the dataset) and the average number of counterfactual points. This is a better suited evaluation metric for our system as it accounts for model performance in the metric of how much further (from the current data distribution) the decision boundary is optimized. The lower the ratio correlates to a better decision boundary of the underlying model. 

The tables following show the results obtained.
\par Description of columns:
\begin{itemize}
    \item Model: Name of the Model (K-Nearest Neighbour (KNN), Neural Network (NN), Random Forest (RF) or Support Vector Machine (SVM))
    \item Counterfactuals (CFs): Number of total counterfactual points generated for queries like `Why P not Q' where P is the predicted class and Q is the desired class ($\neq$ P). Say if a dataset has 3 distinct classes [A,B,C] and the predicted class is A. Then for each datapoint there will be 2 counterfactual options: `Why A not B' \& `Why A not C'.
    \item Common Points (CPs): The number of generated counterfactual points that happen to lie in the dataset.
    \item Ratio: Ratio of number of common points to the total generated counterfactual points.
    \item Average (Avg): Average number of counterfactuals points for the entire dataset for the particular model.
\end{itemize}

The following are the results for the IRIS Dataset: \\

\begin{center}
\begin{tabular}{ | l | l | l | l | l |}
    \hline
    Model & CFs & CPs & Ratio & Avg \\ \hline
    SVM & 472 & 68 & 14.4\% & 7.8 \\ \hline
    RF & 446 & 151 & 33.85\% & 7.4 \\ \hline
    NN & 438 & 67 & 15.29\% & 7.3 \\ \hline
    KNN & 452 & 138 & 30.53\% & 7.5 \\ \hline
\end{tabular}
\end{center}

\par The mobile features dataset is denser than the IRIS dataset. It has more datapoints and substantially more features (21) compared to IRIS(4). The results are as follows: \\

\begin{center}
\begin{tabular}{ | l | l | l | l | l |}
    \hline
    Model & CFs & CPs & Ratio & Avg \\ \hline
    SVM & 18299 & 0 & 0.0\% & 30.4 \\ \hline
    RF & 17619 & 0 & 0.0\% & 29.3 \\ \hline
    NN & 17414 & 0 & 0.0\% & 28.9 \\ \hline
    KNN & 22933 & 30 & 0.1\% & 19.1 \\ \hline
\end{tabular}
\end{center}

\par The results on the Wine Quality Dataset is as follows: \\

\begin{center}
\begin{tabular}{ | l | l | l | l | l |}
    \hline
    Model & CFs & CPs & Ratio & Avg \\ \hline
    SVM & 7018 & 0 & 0.0\% & 4.38 \\ \hline
    RF & 9781 & 209 & 2.13\% & 6.11 \\ \hline
    NN & 8426 & 8 & 0.09\% & 5.26 \\ \hline
    KNN & 6746 & 71 & 1.05\% & 4.21 \\ \hline
\end{tabular}
\end{center}

\subsubsection{Analysis of Results}

We evaluate the results of the pipeline on the basis of the number of common datapoints. The lesser evidently means better. 
As seen, the results on dense datasets (have more features and datapoints) such as the Wine Quality and the Mobile Features data are better (have lesser common points) compared to IRIS. An interesting insight to note is that most of the generated counterfactual points are the ones that are not present in the dataset. Which is, these points would not have been attained on searching the neighborhood space. Independent of the data distribution, our pipeline is able to generated realistic counterfactual point having optimal variations from the target datapoint. 
Comparing the models, we find that SVM and Neural Network are better suited for generating counterfactual points. SVMs are advantageous as they reach the global optimum due to quadratic programming and Neural Networks have the benefit from feature engineering that they are very adept are picking up the feature level nuances. This is sensible given that the whole process and thought behind generating counter factual points is to find the optimum decision boundary separating the desired and the target class. 

\subsection{Systemic Implementation}
To complement the efforts of this paper, we have generated a system \footnote{http://ceh.iiit.ac.in/shap\_dashboard} wherein the user can load either the IRIS or the Mobile Feature dataset and can generate corresponding counterfactual and contrastive explanations. This can later be extended to load any dataset from the user. 

\includegraphics[width=\columnwidth, height=4.0cm]{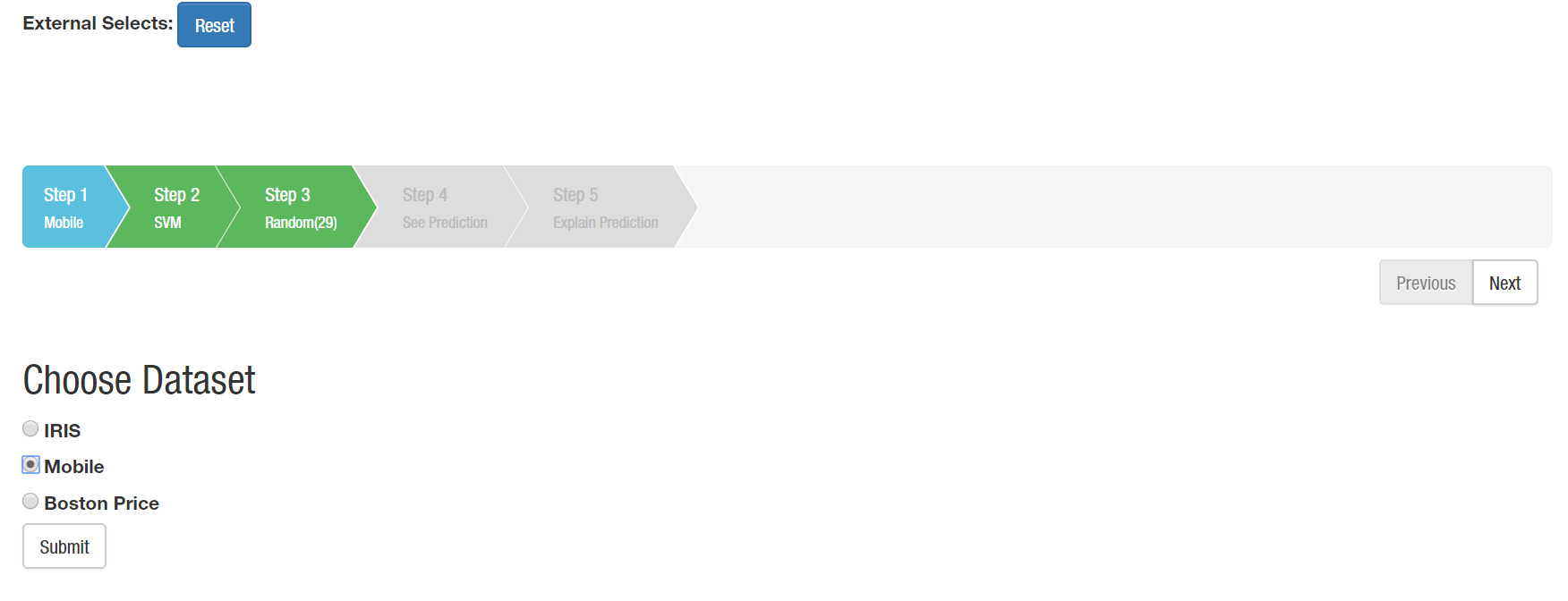}
After selecting the dataset, the user selects the model.
\includegraphics[width=\columnwidth, height=4.0cm]{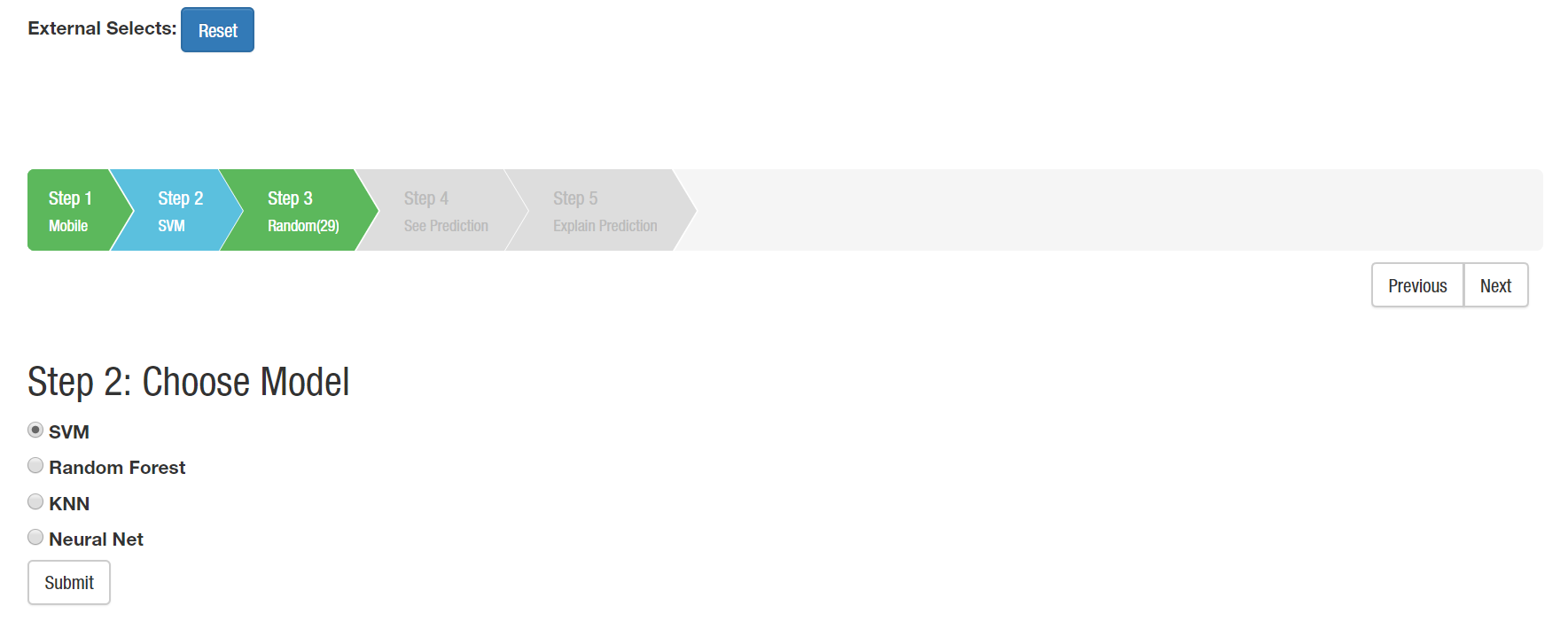}
Post that, based out of a random point from the testing set, the system predicts its class and asks the user for a corresponding Why P not Q type query.
\includegraphics[width=\columnwidth, height=4.0cm]{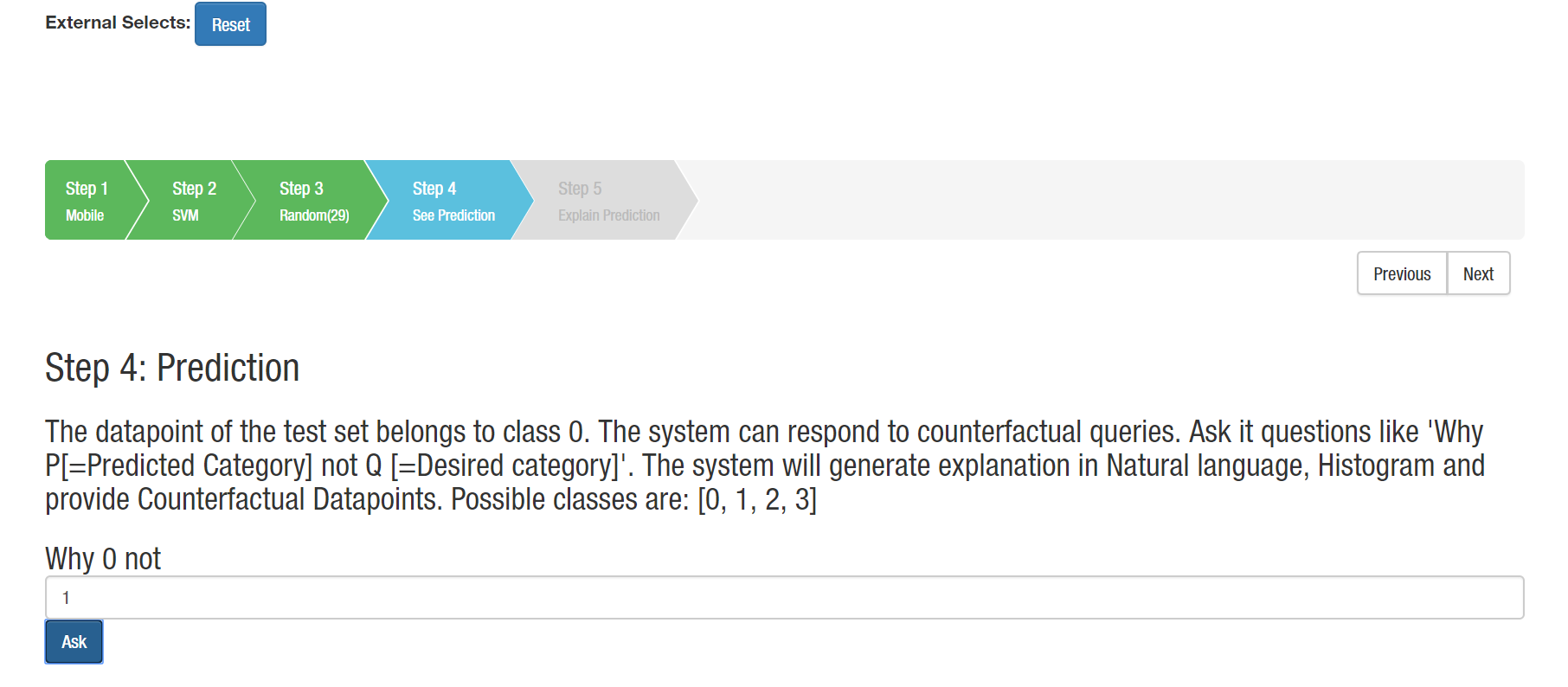}
On this the system generates a contrastive explanation in Natural Language:
\includegraphics[width=\columnwidth, height=4.0cm]{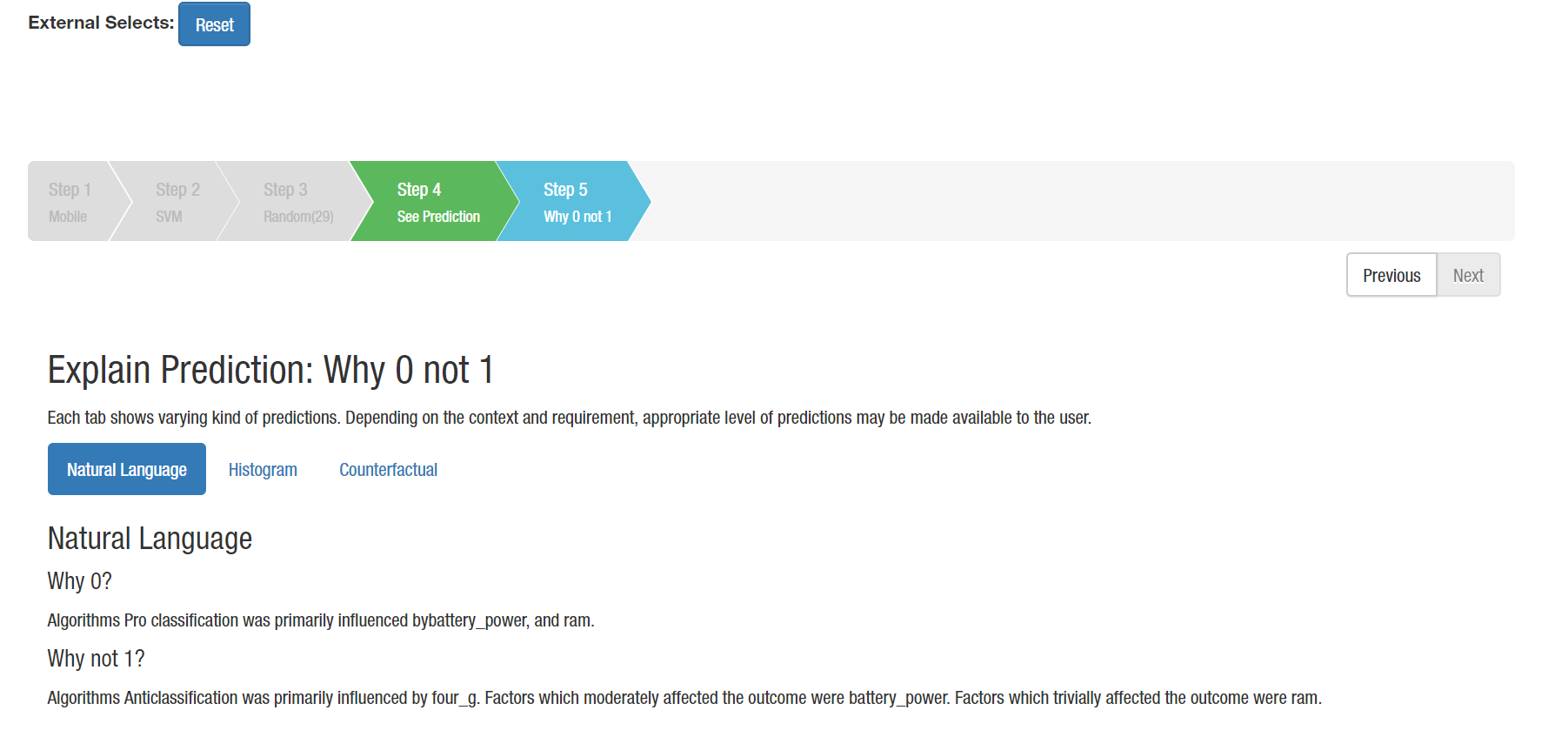}
And, in counterfactual datapoints. Incase if Counterfactual datapoint isnt obtained, we return the nearest desired category datapoint. The image below shows the result of the system generating counterfactual datapoint: \\
\includegraphics[width=\columnwidth, height=4.0cm]{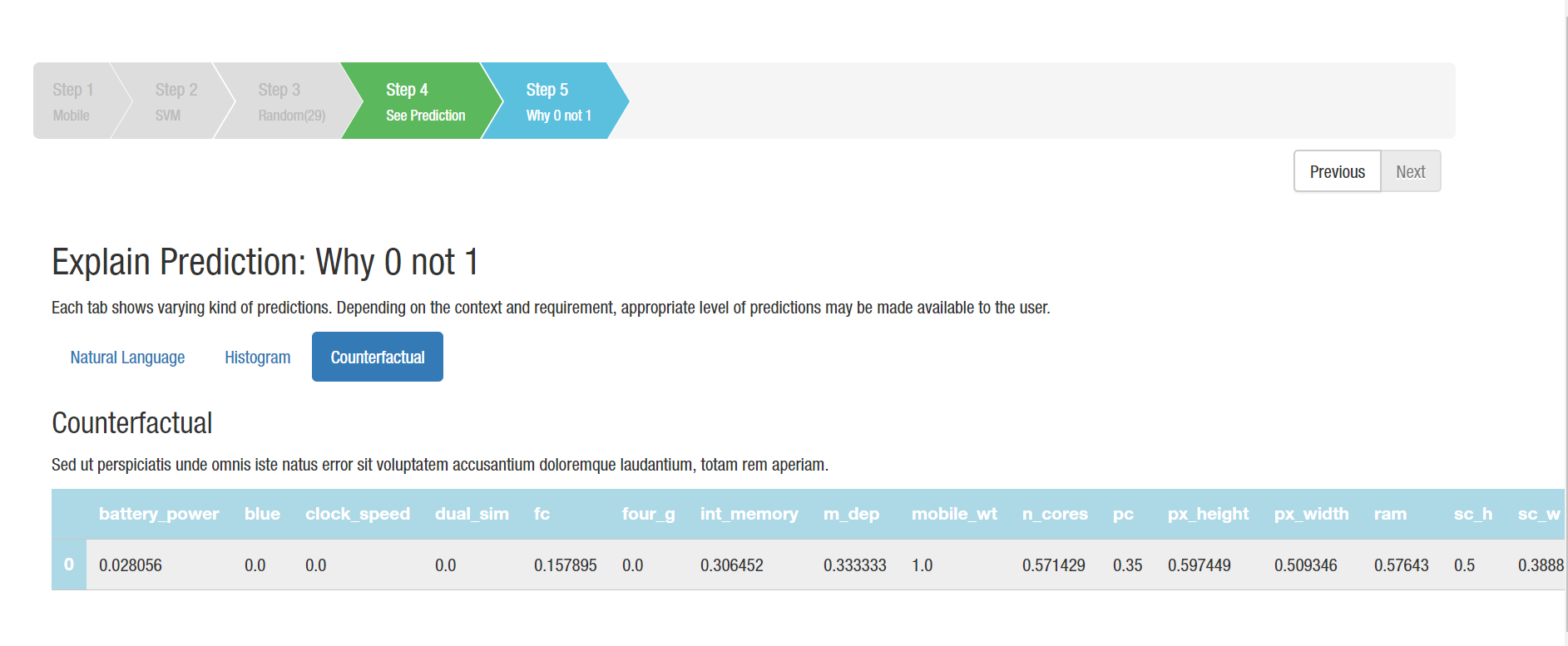}

\section{Conclusion}
This paper builds on the theory of Contrastive and Counterfactual explanation and implements a novel pipeline to generate Contrastive and Counterfactual explanations using Shapley Values. As discussed earlier, Shapley values derived out of Shapley Additive explanations are model agnostic and globally more consistent way of generating explanations.  
This method has a few advantages and a few drawbacks. Besides the above mentioned advantage, another advantage of this method is closely aligns with the concept of counterfactual explanation as we only mutate the features which are adversely impacting our classification task. This is also a possible drawback as the closest counterfactual may not always be near to the mutated set. The most optimum way of generating these counterfactuals would likely be doing adversarial attacks on each feature set and thus is a task of a Generative Adversarial Network (GAN). GANs are efficient but have a huge drawback on performance and time and may not fully be model agnostic. More research has to be led in this direction to understand the the benefits of GANs for the task of generating Counterfactuals. 
\par There is also much improvement possible on a systemic implementation level. The system could be made more universal by addition of features such as support for other datasets. Support for custom models can also be added wherein a trained model could possibly imported as a serialized model object and used in the pipeline. The system can also be designed to be as a conversational agent churning answers to contrastive and counterfactual queries. 
\par This research paves way for development of systems that generate globally consistent contrastive and counterfactual explananda and serves as a base for other more interactive forms of explanation delivery systems.

\bibliographystyle{named}
\bibliography{ijcai19}

\end{document}